\documentclass{article}
\usepackage{ijcai26}

\usepackage{times}
\usepackage{soul}
\usepackage{url}
\usepackage[hidelinks]{hyperref}
\usepackage[utf8]{inputenc}
\usepackage[small]{caption}
\usepackage{graphicx}
\usepackage{amsmath}
\usepackage{amsthm}
\usepackage{booktabs}
\usepackage{algorithm}
\usepackage{algorithmic}
\usepackage[switch]{lineno}
\usepackage{changepage}
\usepackage{multirow}
\usepackage{array}
\usepackage{enumitem}
\usepackage{makecell}
\usepackage{float}
\usepackage{placeins}
\usepackage{CJKutf8}
\usepackage{pifont}
\usepackage[utf8]{inputenc}
\usepackage{tcolorbox}
\tcbuselibrary{skins, breakable}
\usepackage{marvosym}
\definecolor{headerblue}{RGB}{230, 240, 255} 
\definecolor{thoughtgray}{RGB}{248, 248, 248} 
\definecolor{toolblue}{RGB}{240, 248, 255}    
\definecolor{evidcream}{RGB}{255, 253, 245}   
\definecolor{scorehigh}{RGB}{220, 255, 220}  
\definecolor{scoremid}{RGB}{255, 248, 220}   
\definecolor{scorelow}{RGB}{255, 220, 220}   

\newtcolorbox[auto counter, number within=section]{promptbox}[1][]{
  enhanced,
  colback=gray!5,
  colframe=gray!50,
  coltitle=black,
  colbacktitle=gray!25,
  fonttitle=\small\bfseries\scshape,
  toptitle=3pt,
  bottomtitle=3pt,
  fontupper=\small\ttfamily,
  sharp corners,
  boxrule=0.5pt,
  left=5pt, right=5pt, top=5pt, bottom=5pt,
  title={Prompt~\thetcbcounter},
  label={prmt:\thetcbcounter},
  #1
}

\def\@fnsymbol#1{\ensuremath{\ifcase#1\or *\or \dagger\or \ddagger\or
   \mathsection\or \mathparagraph\or \|\or **\or \dagger\dagger
   \or \ddagger\ddagger \else\@ctrerr\fi}}
\newcommand{\ssymbol}[1]{^{\@fnsymbol{#1}}}

\title{RM-Distiller: Exploiting Generative LLM for Reward Model Distillation}

\author{
Hongli Zhou$^{1}$ \and
Hui Huang$^{1}$ \and
Wei Liu$^{1}$ \and
Chenglong Wang$^{2}$ \and
Xingyuan Bu$^{3}$ \and \\
Lvyuan Han$^{1}$ \and
Fuhai Song$^{1}$ \and
Muyun Yang$^{1}$\thanks{Corresponding author.} \and
Wenhao Jiang$^{1}$ \and
Hailong Cao$^{1}$ \And
Tiejun Zhao$^{1}$\\
\affiliations
 \textsuperscript{\rm 1}Faculty of Computing, Harbin Institute of Technology, Harbin, China \\
    \textsuperscript{\rm 2}School of Computer Science and Engineering, Northeastern University, Shenyang, China \\
    \textsuperscript{\rm 3}M-A-P \\
\emails
hongli.joe@stu.hit.edu.cn,
yangmuyun@hit.edu.cn
}

\begin{document}

\maketitle

\begin{CJK*}{UTF8}{gbsn}

\begin{abstract}
Reward models (RMs) play a pivotal role in aligning large language models (LLMs) with human preferences. Due to the difficulty of obtaining high-quality human preference annotations, distilling preferences from generative LLMs has emerged as a standard practice. However, existing approaches predominantly treat teacher models as simple binary annotators, failing to fully exploit the rich knowledge and capabilities for RM distillation. To address this, we propose RM-Distiller, a framework designed to systematically exploit the multifaceted capabilities of teacher LLMs: (1) Refinement capability, which synthesizes highly correlated response pairs to create fine-grained and contrastive signals. (2) Scoring capability, which guides the RM in capturing precise preference strength via a margin-aware optimization objective. (3) Generation capability, which incorporates the teacher's generative distribution to regularize the RM to preserve its fundamental linguistic knowledge. Extensive experiments demonstrate that RM-Distiller significantly outperforms traditional distillation methods both on RM benchmarks and reinforcement learning-based alignment, proving that exploiting multifaceted teacher capabilities is critical for effective reward modeling. To the best of our knowledge, this is the first systematic research on RM distillation from generative LLMs\footnote{Our code is available at \url{https://github.com/Joe-Hall-Lee/RM-Distiller}.}.
\end{abstract}

\section{Introduction}
Large language models (LLMs) have demonstrated remarkable capabilities across diverse domains, and their effective deployment relies on alignment with human values through Reinforcement Learning from Human Feedback (RLHF) \cite{bai2022training}. This paradigm primarily depends on reward modeling, which provides a proxy for human preferences to supervise policy optimization \cite{zhong2025comprehensive}. However, effective reward models (RMs) require training on large-scale, high-quality human preference data. As task complexity scales, obtaining high-quality human annotations becomes prohibitively expensive and difficult~\cite{touvron2023llama}. 

To mitigate reliance on human feedback, some works proposed to directly use powerful generative LLMs as RMs \cite{NEURIPS2023_91f18a12,Huang_He_Zhou_Zhang_Liu_Wang_Liu_Su_2026}. However, despite their strong preference modeling capabilities, their practical use is limited by high API costs and inference latency \cite{liu2025inference}. Consequently, other works have proposed to distill discriminative RMs from generative LLMs~\cite{bai2022constitutional,malik2026rewardbench}. These methods typically employ the LLM as an annotator to assign binary preference labels to response pairs. The resulting preference data with triplets of (\textit{instruction}, \textit{responses}, \textit{label}) are then used to train RMs. Distilling RMs from powerful generative LLMs offers a cost-effective and scalable alternative to human annotation. As a result, it has been adopted widely in recent RL practice \cite{liu2024skywork,liu2025skywork}.

\begin{figure}[t]
    \centering
        \includegraphics[width=0.9\linewidth]{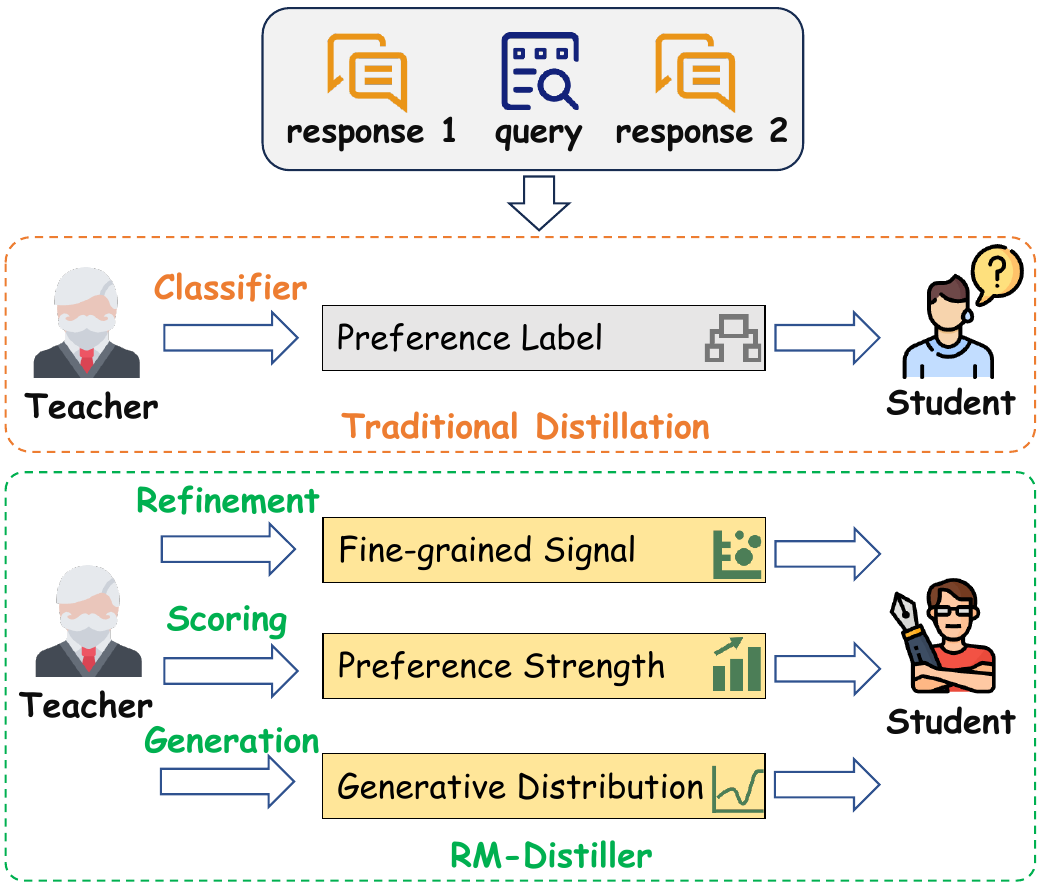}
        \caption{Compared to traditional method, RM-Distiller unlocks the multifaceted potential of teacher LLMs for better distillation.}
    \label{fig:intro}
\end{figure}

Despite its popularity, this paradigm predominantly utilizes the teacher model's capability to assign binary preference labels, often treating it as a mere binary classifier \cite{huang-etal-2025-empirical}, failing to fully leverage the rich capabilities embedded in LLMs. In practice, generative LLMs possess a wide range of capabilities and internal knowledge that are highly beneficial for preference modeling. The insufficient utilization of these capabilities limits the upper bound of RM performance and subsequent RL-based alignment.

We therefore investigate a crucial yet unexplored question: 
\noindent

\begin{adjustwidth}{0.4cm}{0.4cm}
\textit{How can we more effectively distill generative LLMs into reward models?}
\end{adjustwidth}

In this paper, we propose RM-Distiller, a framework that exploits generative LLMs to distill RMs, by incorporating their multifaceted capabilities across three dimensions:

\begin{enumerate}[itemsep=1mm, parsep=0pt]
    \item \textbf{Refinement capability:} We leverage the refinement capability of generative LLMs to construct highly contrastive preference pairs for RM training. By performing targeted revisions on rejected responses, the teacher synthesizes highly correlated response pairs that differ only in the attributes determining preference. This compels the RM to discern subtle quality nuances instead of spurious correlations~\cite{ICLR2025_d41760ec}.
    \item \textbf{Scoring capability:} We utilize the scoring capability of generative LLMs to assess response quality and model preference strength. Instead of relying on discrete binary annotation, the teacher provides continuous preference scores that better reflect varying degrees of quality differences between response pairs. This allows the RM to be optimized with a margin-aware objective, capturing more nuanced preference strengths.
    \item \textbf{Generation capability:} We incorporate the generation capability of generative LLMs to safeguard the foundational linguistic proficiency of the student model. By distilling the teacher's output distribution into the student backbone, we enforce a constraint that prevents the RM from undergoing catastrophic forgetting. This ensures the model retains robustness during regressive RM training \cite{NEURIPS2024_71f71545}.
\end{enumerate}




We evaluate RM-Distiller on both RM benchmarks and downstream RL alignment tasks. The results demonstrate that RM-Distiller significantly outperforms existing distillation methods with superior RM accuracy and RL alignment efficacy. Despite the prevailing practice of distilling generative LLMs into RMs, this is the first work that systematically explores the best practice for RM distillation.

Our contributions are as follows:

\begin{enumerate}[itemsep=1mm, parsep=0pt]
    \item We conduct the first systematic investigation into RM distillation, by introducing a multifaceted distillation strategy RM-Distiller.
    \item We propose to exploit three capabilities of the generative LLMs, \textbf{refinement}, \textbf{scoring}, and \textbf{generation}, to provide comprehensive supervision  for training the RMs.
    \item Extensive experiments demonstrate that RM-Distiller significantly improves RM performance both on RM benchmarks and downstream alignment scenarios.
\end{enumerate}

\section{Background}
\subsection{Preliminary: Reward Modeling}
RMs traditionally relied on human preference datasets, typically trained using the Bradley-Terry (BT) model to predict pairwise rankings \cite{ICLR2025_8e237ec6}. The standard objective of RM training is to learn a scalar function $r_\phi(x, y)$ that reflects human preference. Given a preference dataset $\mathcal{D} = \{(x, y_w, y_l)\}$, where $y_w$ and $y_l$ denote the chosen and rejected responses for instruction $x$, respectively, the RM is optimized by minimizing the negative log-likelihood:

\begin{equation}
\mathcal{L}_\text{BT}(\phi) = - \log \sigma \left( r_\phi(x, y_w) - r_\phi(x, y_l) \right)
\end{equation}
where $\sigma$ is the sigmoid function. While effective for ranking, this loss function only considers the binary order of responses, ignoring the preference strength between them.

\subsection{RM Distillation from Generative LLM}
Due to the high cost and limited scalability of human oversight \cite{touvron2023llama}, the community has increasingly shifted toward getting preference annotation from generative LLM \cite{liu2024skywork}, where teacher models are employed to evaluate paired responses and identify the preferred ones, thereby constructing preference annotations for training BT models \cite{liu2025skywork}. However, this paradigm often treats the teacher as a black-box classifier, failing to exploit its generative potential or the nuanced preference intensities necessary for reliable alignment.

To address this, recent studies have explored ways to enrich the distillation signal by incorporating more diverse information from teacher models. For instance, SynRM \cite{ye-etal-2025-improving} integrates natural language critiques into RMs by appending teacher-generated justifications to the responses during both training and inference. Further extending this idea, CLoud \cite{ankner2024critiqueoutloud} explicitly distills the capability of generating critiques into discriminative RMs. Another line of research focuses on data augmentation, utilizing LLMs to synthesize a larger volume of preference data through response perturbation to cover a broader distribution of scenarios \cite{park-etal-2024-offsetbias,shen2025rmboost}. While these methodologies enhance supervision or data diversity, they often only exploit a fraction of the teacher model's multifaceted capabilities, failing to fully harness its potential for RM distillation.

\section{RM-Distiller}

RM-Distiller begins by Initial Preference Pair Construction, which serves as the foundation for RM distillation. Subsequently, it fully exploits the teacher model's multifaceted capabilities through: (1) Refinement capability for Contrastive Refinement, (2) Scoring capability for Strength Guidance, (3) Generation capability for Generative Regularization.

\begin{figure*}[t]
    \centering
        \includegraphics[width=0.98\linewidth]{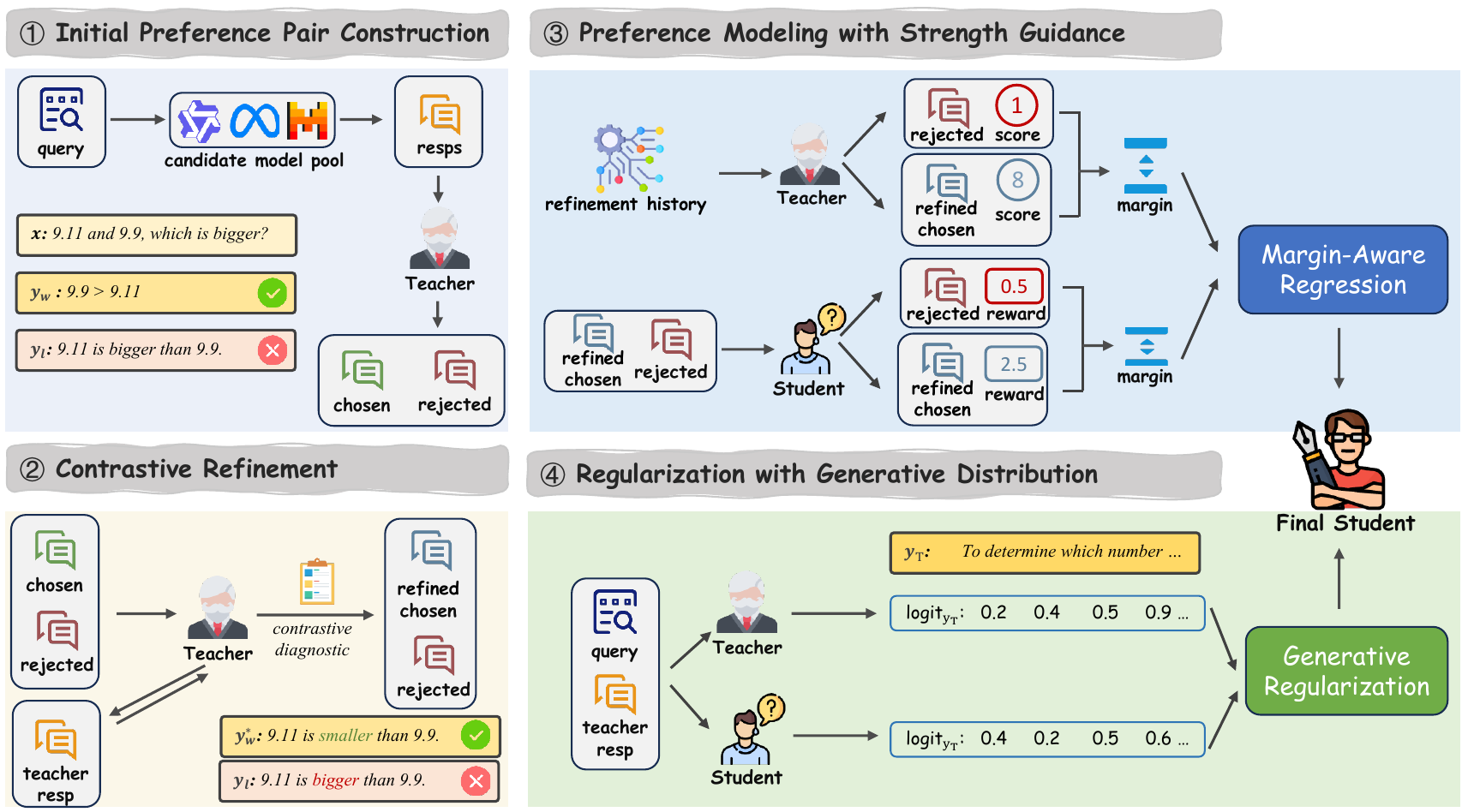}
        \caption{The illustration of RM-Distiller. We first initialize preference pairs from diverse candidate models, and then synthesize highly correlated response pairs. During the training phase, the student model is guided by continuous preference margins and generative regularization.}
    \label{figure:main-fig}
\end{figure*}

\subsection{Initial Preference Pair Construction}
\label{section:init}
\subsubsection{Initial Candidate Generation}

Given an instruction $x$ from an instruction set $\mathcal{D}_x$, we first choose a set of candidate models $\mathcal{M} = \{m_1, m_2, \dots, m_K\}$. After that, we collect a response pool $\mathcal{Y}_x = \{y_1, y_2, \dots, y_K\}$ such that each $y_j \sim m_j(x)$. The candidate models are chosen to differ in model families and sizes, ensuring responses to each instruction are with varying styles and quality levels.


\subsubsection{Initial Preference Annotation}
To construct the initial preference data, for each instruction $x$, we randomly sample two distinct responses $(y_a, y_b)$ from $\mathcal{Y}_x$. We then employ the generative teacher LLM $f_T$ to determine the relative quality between $(y_a, y_b)$. This results in an annotated preference pair $(y_w, y_l)$, where $y_w$ and $y_l$ denote the chosen and rejected responses, respectively.

At this stage, we obtain preference triplets $(x, y_w, y_l)$, which can be directly used to train RMs under the BT objective, as in standard RM distillation. However, this will result in the insufficient utilization of teacher models' capabilities.
Therefore, we introduce the following methods to further leverage the teacher's supervision signal.


\subsection{Contrastive Refinement}
\label{sec:refinement}

The efficacy of RMs depends heavily on the contrastive quality of the preference pairs. In standard distillation practices, pairs are often sampled independently, therefore exhibit significant discrepancies in length, formatting, or linguistic style \cite{ICLR2025_d41760ec}, which results in the RMs to rely on superficial shortcut instead of intrinsic quality attributes \cite{hongli-etal-2024-mitigating}. To eliminate these spurious correlations and force the model to identify fine-grained preference boundaries, we utilize the refinement capability of the teacher model to synthesize highly correlated contrasts, as shown in Figure \ref{contrast-example}.

The refinement process begins by analyzing the gap between the rejected response $y_l$ and the teacher-generated response $y_T$, where $y_T$ serves as a high-quality reference. We first prompt the teacher model $f_T$ to perform a contrastive diagnostic, where it articulates specific reasons for the rejection and outlines a path for improvement. Following this diagnostic, $f_T$ is instructed to minimally modify $y_l$ into a refined chosen version $y_w^*$ that matches the quality of $y_T$ while strictly preserving the original structure and phrasing. By maintaining a minimal edit distance, we ensure that $y_w^*$ and $y_l$ share a nearly identical semantic backbone, differing only in the attributes that define superiority.

\begin{figure}[t]
\centering
\begin{tcolorbox}[
  colback=gray!3,
  colframe=gray!50,
  colbacktitle=gray!15,
  coltitle=black,
  title=Contrastive Refinement Example,
  fonttitle=\bfseries,
  boxrule=0.6pt,
  arc=2mm,
  left=6pt,
  right=6pt,
  top=6pt,
  bottom=6pt,
  titlerule=0.4pt
]
\small

\textbf{Instruction}\\
How to get the last element of a list \texttt{L} in Python?

\vspace{0.4em}
\textbf{Rejected Response ($y_l$)}\\
You can use \textcolor{red!70}{\texttt{L[0]}} to access the last element of a list.

\vspace{0.4em}
\textbf{Refined Chosen Response ($y_w^*$)}\\
You can use \textcolor{green!60!black}{\texttt{L[-1]}} to access the last element of a list.

\vspace{0.4em}
\textbf{Teacher Response ($y_T$)}\\
In Python, negative indexing allows you to access elements from the end, with \texttt{-1} representing the last item.

\end{tcolorbox}
\vspace{-3mm}
\caption{A concrete example of Contrastive Refinement.}
\label{contrast-example}
\end{figure}

\subsection{Preference Modeling with Strength Guidance}
\label{section:scoring}
\subsubsection{Margin-Aware Regression}
In the traditional BT framework, the RM is trained as a binary classifier to maximize the probability $P(y_w \succ y_l)$. This objective implicitly assumes a uniform preference strength across all response pairs. However, in reality, the quality difference between responses varies significantly \cite{wang-etal-2024-reward-modeling}. Therefore, we leverage the scoring capability of LLMs to explicitly capture and represent the strength of preferences to provide fine-grained supervision signals.

First, we prompt the teacher model $f_T$ to assign scalar quality scores $s_T(x, y)$ for chosen and rejected responses $y_w$ and $y_l$, respectively. After that, we propose Margin-Aware Regression, to guide the student model to mimic the teacher's evaluative sensitivity by aligning the predicted reward difference with the teacher-generated score margin. Specifically, we define margin-aware loss as follows:

\begin{equation}
\resizebox{0.91\linewidth}{!}{%
  $\displaystyle
  \mathcal{L}_\text{margin} = \Big( (r_\phi(x, y_w) - r_\phi(x, y_l)) - (s_T(x, y_w) - s_T(x, y_l)) \Big)^2
  $
}
\end{equation}

With this objective, the student model explicitly learns to align not only the relative order but also the exact magnitude of preferences, thereby being able to capture fine-grained variations in preference strength, distinguishing between subtle improvements and absolute errors. While some recent studies have attempted to introduce a margin as a lower bound~\cite{touvron2023llama}, they still fail to capture the nuanced intensity of preference as in this work.

\subsubsection{Self-Calibrated Scoring}
\label{sec:cali}
As established in Section \ref{sec:refinement}, we obtained the refined response $y_w^*$ through Contrastive Refinement as the chosen response.
To assign more precise quality scores, we introduce Self-Calibrated Scoring. Specifically, we explicitly provide the teacher with the previously assigned  score of the rejected response $s_T(x, y_l)$ as a fixed anchor, and task it with scoring the refined response $y_w^*$ on the same scale. As a result, the score $s_T(x, y_w^*)$ reflects a calibrated increase proportional to the specific corrections applied during refinement\footnote{For complete prompts please refer to Appendix \ref{appendix:prompt}.}. This relative evaluation protocol yields more stable and learnable preference margins for downstream reward modeling.

Moreover, during this process, we reuse the refinement dialogue history as the context for scoring, exposing the scorer to the teacher-produced contrastive diagnosis and its refinement outcome. By conditioning the scoring on the explicit refinement history, Self-Calibrated Scoring transforms absolute quality assessment into a relative adjustment.

\subsubsection{Preference Pair Filtering}
\label{sec:filtering}

To ensure high data quality, we filter the synthesized preference pairs to retain sufficiently strong contrasts based on two criteria: (1) a substantial score margin and (2) a minimum edit distance between $y_w^*$ and $y_l$:
\begin{equation}
\begin{aligned}
\mathcal{D}_\text{refine} = \{ &(x, y_w^*, y_l, s_T(x, y_w^*), s_T(x, y_l) \mid  \\
& \text{dist}_\text{edit}(y_w^*, y_l) > \tau_e\land \Delta s > \tau_s \}
\end{aligned}
\end{equation}
where $\tau_s$ and $\tau_e$ denote the minimum thresholds for the score margin and the edit distance, respectively. Filtering for edit distance ensures that the refinement has introduced meaningful changes, while score margin ensures that the teacher recognizes a sufficient quality improvement. This ensures effective contrast between preference pairs for RM training.

\subsection{Regularization with Generative Distribution}
\label{section:generation}
\subsubsection{Generative Regularization}

Solely optimizing for discriminative signals can lead to reward over-optimization, where the model gradually loses its foundational understanding of language, which is essential for generalizability \cite{NEURIPS2024_71f71545}. Therefore, we propose Generative Regularization to preserve the generative knowledge of the backbone model during distillation.

Specifically, we retain the student's language model (LM) head and use the teacher $f_T$ as a linguistic anchor. For each teacher response $y_T$, we supervise the student's LM head using a combination of negative log-likelihood (NLL) and Kullback-Leibler (KL) divergence distillation:

\begin{equation}
\begin{aligned}
\mathcal{L}_{\text{reg}} = \frac{1}{L} \sum_{t=1}^{L} \Big( & -\alpha \cdot \log P_\phi(y^{t}_T | x, y^{<t}_T) \\
& + \beta \cdot \sum_{v \in \mathcal{V}} P_T(v | x, y^{<t}_T) \log \frac{P_T(v | x, y^{<t}_T)}{P_\phi(v | x, y^{<t}_T)} \Big)
\end{aligned}
\end{equation}

\noindent where $y_T = \{y^{1}_T, y^{2}_T, \dots, y^{L}_T\}$ denotes the teacher response of length $L$, and $\mathcal{V}$ denotes the vocabulary. $P_\phi$ and $P_T$ represent the next-token probability distributions of the student and the teacher, respectively. $\alpha$ and $\beta$ are coefficients that control the strength of the NLL loss and the KL divergence. Notably, if the teacher model is closed-source or if the teacher and student do not share a common vocabulary, we set $\beta = 0$. 

\subsubsection{Total Objective Function}
Combining our proposed Margin-Aware Regression and the Generative Regularization loss, the total optimization objective of RM-Distiller is formulated as a weighted sum: 
\begin{equation}
\mathcal{L}_\text{total} = (1 - \alpha - \beta) \cdot \mathcal{L}_\text{margin} + \mathcal{L}_\text{reg}
\end{equation}

This multi-task objective synergizes precise preference modeling with linguistic anchoring. Specifically, $\mathcal{L}_\text{margin}$ enables the model to capture precise quality differences, while $\mathcal{L}_\text{reg}$ prevents representation collapse by grounding the shared backbone in the teacher's generative distribution. This dual supervision ensures semantic robustness, leading to better generalization and stability during downstream RLHF.

\FloatBarrier
\begin{table*}[!ht]
\resizebox{\textwidth}{!}{
\begin{tabular}{lllrrrrrrrrr}
\toprule
\multirow{2}{*}{Student} & \multirow{2}{*}{Teacher} & \multirow{2}{*}{Method}
& \multicolumn{5}{c}{RewardBench} & \multicolumn{4}{c}{RM-Bench} \\
\cmidrule(lr){4-8} \cmidrule(lr){9-12}
& & & Chat & Hard & Safety & Reason & Avg.
& Easy & Normal & Hard & Avg. \\

\midrule
\multicolumn{12}{l}{\textit{LLM-as-a-Judge}} \\
\midrule

\multicolumn{3}{l}{Claude-3.5-Sonnet-20240620} 
& 96.4 & 74.0 & 81.6 & 84.7 & 84.2 & 73.8 & 63.4 & 45.9 & 61.0 \\
\multicolumn{3}{l}{Llama-3.1-70B-Instruct \cite{grattafiori2024llama}} 
& 97.2 & 70.2 & 82.8 & 86.0 & 84.1 & 74.7 & 67.8 & 54.1 & 65.5 \\
\multicolumn{3}{l}{GPT-4o \cite{hurst2024gpt}}
& 96.1 & 70.7 & 87.8 & 75.1 & 82.4 & 83.8 & 70.9 & 48.2 & 67.6 \\
\multicolumn{3}{l}{Qwen3-14B \cite{yang2025qwen3}}
& 96.4 & 69.2 & 88.0 & 82.0 & 83.9 & 83.6 & 71.7 & 52.1 & 69.1 \\
\multicolumn{3}{l}{Qwen2.5-3B-Instruct \cite{qwen2024qwen25}}
& 88.3 & 43.2 & 55.1 & 42.7 & 57.3 & 69.6 & 49.4 & 34.3 & 51.1 \\

\midrule
\multicolumn{12}{l}{\textit{Open-Source Reward Models}} \\
\midrule

\multicolumn{3}{l}{Eurus-RM-7B \cite{ICLR2025_3e2c12c1}}
& 98.0 & 65.6 & 81.4 & 86.3 & 81.6 & 87.2 & 70.2 & 40.2 & 65.9 \\

\multicolumn{3}{l}{GRAM-LLaMA3.2-3B-RewardModel \cite{pmlr-v267-wang25ad}} 
& 90.6 & 83.9 & 79.8 & 80.2 & 83.6 & 84.2 & 65.2 & 40.0 & 63.2 \\
\multicolumn{3}{l}{JudgeLRM-7B \cite{chen2025judgelrm}} 
& 92.9 & 56.4 & 78.2 & 73.6 & 75.2 & 73.2 & 66.2 & 54.8 & 64.7 \\

\midrule

\multicolumn{3}{l}{\textit{Training on the Same Preference Dataset in the Unlabeled Scenario}} \\
\midrule
\multirow{16}{*}{Qwen2.5-3B-Instruct} & \multirow{9}{*}{GPT-4o} & BT Classifier
& 93.3 & 37.7 & 69.3 & 81.0 & 70.3
& \textbf{91.9} & 62.3 & 13.6 & 55.9 \\

& & Margin BT \cite{touvron2023llama}
& 93.6 & 33.6 & 73.5 & 80.8 & 70.4
& 90.4 & 61.0 & 17.5 & 56.3 \\

& & SteerLM \cite{dong-etal-2023-steerlm}
& 94.4 & 46.5 & 74.3 & 82.8 & 74.5
& 90.2 & 59.8 & 17.1 & 55.7 \\

& & SynRM \cite{ye-etal-2025-improving}
& \textbf{96.1} & 47.6 & 78.7 & 79.1 & 75.4
& 91.0 & \textbf{66.2} & 25.1 & 60.7 \\

& & RMBoost \cite{shen2025rmboost}
& 91.3 & 35.1 & 67.8 & 74.8 & 67.3
& 86.3 & 54.9 & 17.6 & 52.9 \\

& & CLoud \cite{ankner2024critiqueoutloud}
& 93.6 & 37.1 & 62.6 & 63.6 & 64.2
& 87.1 & 59.2 & 18.9 & 55.1 \\

& & LSAM \cite{wang-etal-2024-reward-modeling}
& 94.1 & 37.3 & 73.5 & 81.4 & 71.6
& 89.3 & 58.8 & 22.2 & 56.8 \\

& & \textbf{RM-Distiller}
& 95.5 & \textbf{51.1} & \textbf{84.1} & \textbf{87.4} & \textbf{79.5}
& 86.4 & 64.8 & \textbf{32.2} & \textbf{61.1} \\

\cmidrule{2-12}
& \multirow{8}{*}{Qwen3-14B} & BT Classifier
& 88.0 & 43.4 & 72.4 & 83.5 & 71.8
& 89.8 & 60.7 & 19.8 & 56.7 \\

& & Margin BT \cite{touvron2023llama}
& 94.7 & 46.3 & 79.9 & 70.4 & 72.8
& 90.7 & 63.6 & 26.3 & 60.2 \\

& & SteerLM \cite{dong-etal-2023-steerlm}
& 93.6 & 45.0 & 79.1 & 78.5 & 74.0
& 89.9 & 64.1 & 19.9 & 58.0 \\

& & SynRM \cite{ye-etal-2025-improving}
& 84.9 & \textbf{57.0} & 79.1 & \textbf{91.1} & 78.0
& \textbf{92.7} & \textbf{66.7} & 20.0 & 59.8 \\

& & RMBoost \cite{shen2025rmboost}
& 92.7 & 43.6 & 73.7 & 90.6 & 75.2
& 84.8 & 60.3 & 25.9 & 57.0 \\

& & CLoud \cite{ankner2024critiqueoutloud}
& 91.1 & 39.7 & 74.1 & 67.5 & 68.1
& 87.7 & 58.4 & 20.3 & 55.5 \\

& & LSAM \cite{wang-etal-2024-reward-modeling}
& 92.5 & 39.9 & 72.7 & 84.9 & 72.5
& 88.8 & 60.4 & 24.6 & 57.9 \\

& & \textbf{RM-Distiller}
& \textbf{95.0} & 55.0 & \textbf{81.8} & 89.0 & \textbf{80.2}
& 87.2 & 66.3 & \textbf{34.6} & \textbf{62.7} \\
\midrule
\multicolumn{3}{l}{\textit{Training on the Same Preference Dataset in the Labeled Scenario}} \\
\midrule
\multirow{15}{*}{Qwen2.5-3B-Instruct} & --- & BT Classifier
& 77.4 & 71.3 & 86.5 & 60.0 & 73.8
& 81.7 & 64.5 & 43.2 & 63.2 \\
\cmidrule{2-12}
& \multirow{7}{*}{GPT-4o} & Margin BT \cite{touvron2023llama}
& 88.6 & 70.8 & 87.3 & 67.8 & 78.6
& 80.3 & 66.8 & \textbf{48.2} & 65.1 \\

& & SteerLM \cite{dong-etal-2023-steerlm}
& 91.6 & 65.4 & 80.4 & 63.5 & 75.2
& 85.6 & 65.8 & 37.4 & 62.9 \\

& & SynRM \cite{ye-etal-2025-improving}
& 81.0 & \textbf{82.5} & 86.9 & 78.6 & 82.2
& 80.9 & 66.0 & 40.3 & 62.4 \\

& & RMBoost \cite{shen2025rmboost}
& 85.8 & 73.7 & 86.6 & 74.9 & 80.3
& 79.3 & 59.0 & 38.8 & 59.1 \\

& & CLoud \cite{ankner2024critiqueoutloud}
& 69.3 & 66.0 & 72.6 & 66.9 & 68.7
& 68.3 & 57.1 & 45.4 & 57.0 \\

& & LSAM \cite{wang-etal-2024-reward-modeling}
& 89.4 & 66.5 & 88.1 & 63.8 & 77.0
& 85.2 & 63.8 & 36.7 & 61.9 \\

& & \textbf{RM-Distiller}
& \textbf{92.7} & 64.3 & \textbf{90.7} & \textbf{86.9} & \textbf{83.6}
& \textbf{87.0} & \textbf{68.9} & 42.0 & \textbf{65.9} \\
\cmidrule{2-12}
& \multirow{7}{*}{Qwen3-14B} & Margin BT \cite{touvron2023llama}
& 88.6 & 72.8 & 86.6 & 71.6 & 79.9
& 81.0 & 67.8 & \textbf{49.1} & 66.0 \\

& & SteerLM \cite{dong-etal-2023-steerlm}
& 89.7 & 69.5 & \textbf{87.2} & 67.1 & 78.4
& 84.6 & 68.3 & 43.8 & 65.6 \\

& & SynRM \cite{ye-etal-2025-improving}
& 78.2 & \textbf{80.7} & 85.1 & 74.8 & 79.7
& 73.6 & 61.5 & 48.5 & 61.2 \\

& & RMBoost \cite{shen2025rmboost}
& 82.7 & 73.7 & 82.2 & 88.2 & 81.7
& 84.8 & 60.3 & 25.9 & 57.0 \\

& & CLoud \cite{ankner2024critiqueoutloud}
& 83.0 & 67.3 & 75.0 & 62.2 & 71.9
& 77.3 & 61.9 & 44.9 & 61.4 \\

& & LSAM \cite{wang-etal-2024-reward-modeling}
& \textbf{93.0} & 69.7 & 84.1 & 70.4 & 79.3
& \textbf{90.7} & 66.0 & 35.2 & 63.9 \\

& & \textbf{RM-Distiller}
& 91.3 & 69.7 & 85.5 & \textbf{91.3} & \textbf{84.5}
& 89.5 & \textbf{69.3} & 40.5 & \textbf{66.4} \\
\bottomrule
\end{tabular}
}
\centering
\caption{Results of different RM distillation methods on RewardBench and RM-Bench.}
\label{tab:main_results}
\end{table*}

\section{Experiments}
In this section, we evaluate RM-Distiller through a series of experiments to assess its reward modeling performance.

\subsection{Experimental Setup}

\subsubsection{Data Settings}

For training RMs, we utilize a 10,000-sample subset randomly drawn from the Skywork-Preference-v0.2 dataset \cite{liu2024skywork}. Our method is evaluated across two  scenarios:

\begin{itemize}[leftmargin=*]
    \item \textbf{Unlabeled Scenario} where no prior preference annotations are available, and we only utilize instructions from the dataset and employ candidate models to generate responses. This setup mirrors practical distillation settings with no access to high-quality human annotations, making LLM-based distillation a more practical alternative.
    \item \textbf{Labeled Scenario} where response pairs are associated with high-quality ground-truth preference annotations, and we retain the original response pairs from the dataset and keep only those samples for which the teacher's preference judgments are consistent with the original labels.
\end{itemize}

We employ RewardBench \cite{lambert-etal-2025-rewardbench} and RM-Bench \cite{ICLR2025_6da1eec8} as our primary evaluation benchmarks, which cover diverse subsets.

\subsubsection{Models}

We employ Qwen2.5-3B-Instruct \cite{qwen2024qwen25} as the backbone for the student RM. To facilitate distillation, we use GPT-4o \cite{hurst2024gpt} and Qwen3-14B \cite{yang2025qwen3} in its non-thinking mode as the teacher models. For constructing the response pool in the unlabeled scenario, we leverage a set of candidate models including Llama-3.1-8B-Instruct \cite{grattafiori2024llama}, Mistral-7B-Instruct-v0.3 \cite{jiang2023mistral7b}, Gemma-2-9b-it \cite{team2024gemma} and Vicuna-7b-v1.5 \cite{NEURIPS2023_91f18a12}.

\begin{table*}[!t]
\centering
\resizebox{\textwidth}{!}{
\begin{tabular}{llllllrrrrrrr}
\toprule
\multirow{2}{*}{Policy}
& \multirow{2}{*}{Student RM}
& \multirow{2}{*}{Teacher}
& \multirow{2}{*}{Dataset}
& \multirow{2}{*}{Method}
& \multirow{2}{*}{Algorithm}
& \multicolumn{2}{c}{AlpacaEval 2.0}
& \multicolumn{2}{c}{FollowBench}
& \multicolumn{3}{c}{CFBench} \\
\cmidrule(lr){7-8} \cmidrule(lr){9-10} \cmidrule(lr){11-13}
& & & & & 
& LC & Len
& HSR & SSR
& CSR & ISR & PSR \\
\midrule
\multirow{13}{*}{Llama-3-8B}
& —
& —
& UltraChat
& Baseline
& SFT
& 8.86 & 1,017
& 41.04 & 57.39
& 0.51 & 0.15 & 0.22 \\
\cmidrule(lr){2-13}
& \multirow{12}{*}{\shortstack{Qwen2.5-3B\\-Instruct}}
& \multirow{12}{*}{GPT-4o}
& \multirow{6}{*}{HH-RLHF}
& \multirow{3}{*}{BT Classifier}
& PPO   & 14.32 & 2,045 & 46.85 & 61.20 & 0.57 & 0.18 & 0.26 \\
& & & &  & GRPO  & 17.89 & 2,310 & 48.33 & 62.15 & 0.61 & 0.22 & 0.30 \\
& & & &  & DAPO  & 18.25 & 2,005 & 48.59 & 62.48 & 0.62 & 0.22 & 0.31 \\
\cmidrule(lr){5-13}
& & & & \multirow{3}{*}{RM-Distiller}
& PPO   & 16.90 & 1,927 & 48.42 & 62.05 & 0.60 & 0.23 & 0.29 \\
& & & &  & GRPO  & 20.10 & 2,342 & 49.95 & 63.14 & \textbf{0.64} & \textbf{0.24} & \textbf{0.33} \\
& & & &  & DAPO  & \textbf{20.52} & 2,487 & \textbf{50.09} & \textbf{63.31} & 0.63 & 0.23 & 0.32 \\
\cmidrule(lr){4-13}
& & & \multirow{6}{*}{ShareGPT}
& \multirow{3}{*}{BT Classifier}
& PPO   & 13.80 & 2,584 & 46.20 & 60.95 & 0.56 & 0.17 & 0.25 \\
& & & &  & GRPO  & 17.92 & 2,360 & 47.90 & 61.88 & 0.61 & 0.21 & 0.30 \\
& & & &  & DAPO  & 17.45 & 2,200 & 48.15 & 62.10 & 0.60 & 0.21 & 0.29 \\
\cmidrule(lr){5-13}
& & & & \multirow{3}{*}{RM-Distiller}
& PPO   & 16.24 & 2,293 & 47.78 & 61.77 & 0.59 & 0.19 & 0.27 \\
& & & &  & GRPO  & 19.85 & 2,090 & 49.45 & 62.90 & 0.62 & 0.22 & 0.31 \\
& & & &  & DAPO  & \textbf{20.15} & 2,335 & \textbf{49.62} & \textbf{63.15} & \textbf{0.63} & \textbf{0.23} & \textbf{0.32} \\
\bottomrule
\end{tabular}
}
\caption{Performance of policy model supervised by different RM distillation methods.}
\label{tab:rl_results}
\end{table*}

\begin{table}[t]
\centering
\setlength{\tabcolsep}{4pt}
\resizebox{0.95\columnwidth}{!}{
\begin{tabular}{lrr}
\toprule
\multirow{2}{*}{Method} & RewardBench & RM-Bench \\
& Average & Average \\
\midrule
\multicolumn{3}{l}{\textit{GPT-4o as Teacher and Qwen2.5-3B-Instruct as Student}} \\
\midrule
BT Classifier                & 70.3 & 55.9 \\
+ Margin-Aware Regression    & 72.7 & 57.3 \\
+ Contrastive Refinement     & 76.8 & 58.8 \\
+ Generative Regularization  & \textbf{79.5} & \textbf{61.1} \\
\midrule
\multicolumn{3}{l}{\textit{Qwen3-14B as Teacher and Qwen2.5-3B-Instruct as Student}} \\
\midrule
BT Classifier                & 71.8 & 56.7 \\
+ Margin-Aware Regression    & 75.7 & 59.5 \\
+ Contrastive Refinement     & 78.7 & 61.5 \\
+ Generative Regularization  & \textbf{80.2} & \textbf{62.7} \\
\bottomrule
\end{tabular}
}
\caption{Ablation study of RM-Distiller components.}
\label{tab:ablation}
\end{table}

\subsubsection{Baselines}
\label{section:baselines}

We compare RM-Distiller against several RM distillation methods including the vanilla BT Classifier, alongside its extensions Margin BT \cite{touvron2023llama} and LSAM \cite{wang-etal-2024-reward-modeling}. We also evaluate against methods that leverage teacher-generated critiques, revisions or attributes, including SynRM \cite{ye-etal-2025-improving}, CLoud \cite{ankner2024critiqueoutloud}, RMBoost \cite{shen2025rmboost}, and SteerLM \cite{dong-etal-2023-steerlm}\footnote{For more details please refer to Appendix \ref{appendix:imple_detail}.}.

We also include a range of external RMs as reference points. These comprise discriminative models such as Eurus-RM-7B \cite{ICLR2025_3e2c12c1}, as well as off-the-shelf generative models including Claude-3.5-Sonnet and Llama-3.1-70B-Instruct, along with fine-tuned generative models including GRAM-LLaMA3.2-3B-RewardModel \cite{pmlr-v267-wang25ad} and JudgeLRM-7B \cite{chen2025judgelrm}.

\subsection{Main Results}
\label{sec:main}

As shown in Table \ref{tab:main_results}, RM-Distiller consistently and significantly outperforms traditional distillation methods across both unlabeled and labeled settings, whether using a closed-source or open-source model as a teacher. Notably, RM-Distiller also surpasses several powerful external baselines, including both frontier generative models and open-source RMs built on significantly larger foundation models or trained with more extensive datasets.

These performance gains are driven by the effective synergy of three key capabilities leveraged from the teacher model. Specifically, the teacher's refinement and scoring abilities enable the RM to identify fine-grained differences in response pairs, thereby enhancing the perceptual capability of nuanced quality aspects. Furthermore, the teacher's generation capability helps the RM maintain a foundational understanding of language, allowing it to better comprehend what constitutes a good response.

\subsection{Ablation Study}

To further verify the effectiveness of RM-Distiller, we conduct an ablation study by incrementally adding each module, corresponds to each capability of the teacher. As shown in Table \ref{tab:ablation}, each proposed module contributes significantly to the overall performance gains. Across both RewardBench and RM-Bench, we observe a steady increase in average accuracy as we incrementally integrate Margin-Aware Regression, Contrastive Refinement, and Generative Regularization. This confirms that leveraging the teacher's multifaceted capabilities is essential for high-fidelity RM distillation on both evaluation benchmarks\footnote{Additional ablation studies are reported in Appendix~\ref{appendix:experiment}.}.

\subsection{Performance on RLHF}
\label{sec:rlhf}

To evaluate the effectiveness of the distilled RMs in downstream alignment, we conduct RLHF experiments using PPO \cite{schulman2017proximal}, GRPO \cite{shao2024deepseekmath}, and DAPO \cite{NEURIPS2025_a4277440}. For all RL experiments, we adopt Llama-3-8B \cite{grattafiori2024llama} as the base policy model. We first perform SFT on the base model using the UltraChat dataset \cite{ding-etal-2023-enhancing} as a cold-start. We then use prompts from the HH-RLHF \cite{bai2022training} and ShareGPT \cite{NEURIPS2023_91f18a12} datasets as training instructions during RL. Alignment effectiveness are evaluated on both general and complex instruction following benchmarks, including AlpacaEval 2.0 \cite{dubois2024lengthcontrolled}, FollowBench \cite{jiang-etal-2024-followbench} and CFBench \cite{zhang-etal-2025-cfbench}\footnote{Since these benchmarks rely on LLM-based evaluation, we adopt GPT-4o-0806 as the judge model.}.

As shown in Table~\ref{tab:rl_results}, RM-Distiller consistently outperforms the BT Classifier across all datasets and RL algorithms, indicating more effective reward signal for RLHF. Our distilled RM significantly enhances both general alignment and complex instruction following, demonstrating its ability to discern nuanced quality differences. Furthermore, the consistent gains under length-controlled evaluations confirm that these improvements reflect genuine enhancements in response quality rather than a mere verbosity bias.


\begin{table}[t]

\resizebox{\linewidth}{!}{
\begin{tabular}{lrr}
\toprule
Model & Sample Num & Acc. \\
\midrule
Skywork-Reward-V2-8B \cite{liu2025skywork}   & 40,000K & 81.3 \\
Skywork-Reward-8B-v0.2 \cite{liu2024skywork} & 80K     & 75.9 \\
URM-LLaMA-3.1-8B \cite{lou2024uncertainty}      & 100K    & 76.7 \\
Llama-3-OffsetBias-8B \cite{park-etal-2024-offsetbias}          & 70K     & 83.2 \\
Tulu-3-8B-RM-RB2 \cite{malik2026rewardbench}     & 350K    & 83.2 \\
\midrule
BT-Qwen2.5-3B-Instruct            & 10K     & 72.2 \\
\textbf{RM-Distiller-Qwen2.5-3B-Instruct}  & 10K     & \textbf{85.4} \\
\bottomrule
\end{tabular}
}
\centering
\caption{Results of different methods and previous RMs on the Arabic preference test set.}
\label{tab:adaptaion}
\end{table}

\begin{table}[t]
\centering
\resizebox{\linewidth}{!}{
\begin{tabular}{lllr}
\toprule
Student & Teacher & Method & EvalBiasBench \\

\midrule
\multicolumn{3}{l}{GPT-4o \cite{hurst2024gpt}} & 70.0 \\
\multicolumn{3}{l}{Qwen3-14B \cite{yang2025qwen3}} & 75.0 \\
\multicolumn{3}{l}{Qwen2.5-3B-Instruct \cite{qwen2024qwen25}} & 44.4 \\
\midrule
\multirow{4}{*}{Qwen2.5-3B-Instruct} & \multirow{2}{*}{GPT-4o} & BT Classifier & 30.0 \\
& & \textbf{RM-Distiller} & \textbf{57.5} \\
\cmidrule{2-4}
& \multirow{2}{*}{Qwen3-14B} & BT Classifier & 47.5 \\ 
& & \textbf{RM-Distiller} & \textbf{63.8} \\
\bottomrule
\end{tabular}
}
\caption{Results of different methods on EvalBiasBench.}
\label{tab:evalbiasbench}
\end{table}

\begin{table}[t]
\centering
\resizebox{\linewidth}{!}{
\begin{tabular}{lllr}
\toprule
Student & Teacher & Method & IFBench \\
\midrule
\multicolumn{3}{l}{GPT-4o \cite{hurst2024gpt}} & 50.8 \\
\multicolumn{3}{l}{Qwen3-14B \cite{yang2025qwen3}} & 54.2 \\
\multicolumn{3}{l}{Qwen2.5-3B-Instruct \cite{qwen2024qwen25}} & 36.9 \\
\midrule
\multirow{4}{*}{Qwen2.5-3B-Instruct} & \multirow{2}{*}{GPT-4o} & BT Classifier & 50.2 \\
& & \textbf{RM-Distiller} & \textbf{57.9} \\
\cmidrule{2-4}
& \multirow{2}{*}{Qwen3-14B} & BT Classifier & 52.5 \\
& & \textbf{RM-Distiller} & \textbf{57.7} \\
\bottomrule
\end{tabular}
}
\caption{Results of different methods on IFBench.}
\label{tab:ifbench}
\end{table}

\subsection{Fast Adaptation Capability}

Fast Adaptation refers to the ability to rapidly provide reliable reward signals for novel, domain-specific tasks, which is a vital determinant of RM's efficacy. RM-Distiller offers an efficient and practical solution, enabling the rapid distillation of specialized RMs with only a small amount of targeted data. To evaluate this capability, we choose the Arabic language domain, where we utilize an Arabic dataset\footnote{\url{https://huggingface.co/datasets/FreedomIntelligence/Arabic-preference-data-RLHF}}, which has significant difference with general preference modeling\footnote{We randomly sample 10,000 instructions as the start point for distillation. For evaluation, we extract a test set of 1,000 samples, which are filtered by GPT-5 to ensure a reliable ground truth.}. We utilize Qwen2.5-3B-Instruct as the student model and Qwen3-14B as the teacher model.

As shown in Table \ref{tab:adaptaion}, RM-Distiller not only demonstrates a substantial margin over the BT model but also surpasses currently top-performing generalist RMs. This empirical evidence reveals that generalist RMs struggle to achieve true universality across diverse specific domains. In contrast, RM-Distiller provides an efficient mechanism to adapt and distill reliable reward signals into specialized scenarios with only a small group of instruction data, enabling fast adaptation of reward modeling to a new domain.

\begin{table}[t]
\centering
\small
\setlength{\tabcolsep}{4pt}
\begin{tabular}{llrr}
\toprule
\multirow{2}{*}{Teacher} & \multirow{2}{*}{Method} & RewardBench & RM-Bench \\
& & Average & Average \\
\midrule
\multirow{2}{*}{GPT-4o}
& BT Classifier   & 50.3 & 48.1 \\
& \textbf{RM-Distiller} & \textbf{76.1} & \textbf{59.9} \\
\midrule
\multirow{2}{*}{Qwen3-14B}
& BT Classifier   & 64.6 & 51.9 \\
& \textbf{RM-Distiller} & \textbf{76.4} & \textbf{60.2} \\
\bottomrule
\end{tabular}
\caption{Results of different methods under 1K instructions.}
\label{tab:data_efficiency}
\end{table}



\subsection{Generalizability}

A robust RM should generalize to challenging scenarios without exploiting spurious shortcuts. Poor generalizability often leads to reward hacking, where the model exploits unintended biases to achieve deceptively high scores without fully following the instructions \cite{zhou2026toward}. In this section, we employ EvalBiasBench \cite{park-etal-2024-offsetbias} and IFBench \cite{peng-etal-2025-agentic} as our evaluation benchmarks.

As shown in Table~\ref{tab:evalbiasbench} and \ref{tab:ifbench}, RM-Distiller demonstrates a superior capacity to generalize to out-of-distribution tasks. In EvalBiasBench, while the BT Classifier collapses by relying on superficial patterns, RM-Distiller exhibits significantly higher robustness. This indicates the student model learns to discern semantic boundaries rather than spurious correlations. In IFBench, RM-Distiller not only outperforms the BT Classifier baseline but even surpasses the direct prompting performance of the teacher. This demonstrates that RM-Distiller effectively captures fine-grained constraints within instructions, verifying its generalizability to unseen tasks.

\subsection{Data Efficiency}

Data efficiency is critical in RM distillation, as high-quality instruction data is often scarce when dealing with specialized domains or rapid model iteration. To evaluate this, we compare RM-Distiller against the BT baseline using a reduced training set of only 1,000 instructions.

As illustrated in Table \ref{tab:data_efficiency}, when the training set is reduced to only 1,000 instructions, the performance of the student model trained with BT model deteriorates sharply. In contrast, RM-Distiller exhibits exceptional data efficiency, maintaining a high level of discriminative accuracy. By extracting significantly more supervisory information from each sample, RM-Distiller effectively distills the teacher's underlying preferences. This high-density training signal enables the development of reliable RMs with minimal data, offering a practical solution for real-world alignment tasks.

\section{Conclusion}
In this work, we present the first systematic research about how to distill generative LLMs into discriminative RMs. By exploiting the multifaceted capabilities of generative LLMs, our proposed RM-Distiller provides the student model with rich, reliable supervision signals, enabling high RM performance without additional inference overhead.

Given the rapid progress of frontier LLMs, alignment can be done from both human and AI feedback. In the future, we will try to distill preference signals from human-LLM collaboration to further improve alignment efficacy.

\section*{Acknowledgments}
This work was supported by the National Key R\&D Program of China (Grant No. 2024YFC3809100), the National Natural Science Foundation of China (Grant No. 62276077, 62376075, 62376076), the Department of Science and Technology of Heilongjiang Province (Grant No. ZL2025F004), and the MOE Key Laboratory of Cognitive Intelligence and Content Security (Grant No. 10120251103, Harbin Institute of Technology).

\section*{Contribution Statement}
Hongli Zhou, Hui Huang and Wei Liu contributed equally.

\bibliographystyle{named}
\bibliography{ijcai26}

\clearpage

\appendix

\section{Implementation Details}
\label{appendix:imple_detail}

\subsection{Training Details}

For RM-Distiller training, as detailed in Section \ref{sec:refinement}, we use Contrastive Refinement to synthesize highly correlated response pairs. To enhance the richness of the training signals, we construct the final training set $\mathcal{D}_\text{final}$ by merging the teacher-refined data with the original sampled preference pairs, such that $\mathcal{D}_\text{final} = \mathcal{D}_\text{sample} \cup \mathcal{D}_\text{refine}$. We perform training on 2 NVIDIA A800 GPUs, and the training requires approximately 3 GPU hours. The specific hyperparameters for RM training are shown in Table \ref{tab:hyperparams_rm}.

For RL experiments in Section \ref{sec:rlhf}, we use 16 H20 NVIDIA GPUs. The specific hyperparameters for RL are shown in Table \ref{tab:hyperparams_rl}.

\subsection{Inference Details}
During the response generation phase in Section \ref{sec:main}, we ensure consistency in sampling by setting the temperature to 0 and the top-p value to 1 for most candidate models. For the teacher model Qwen3-14B, we follow the official recommendations for its non-thinking mode\footnote{\url{https://huggingface.co/Qwen/Qwen3-14B}}. Specifically, we set the sampling parameters with a temperature of 0.7, a top-p value of 0.8, a top-k threshold of 20, and a min-p value of 0.

\subsection{Baseline Details}
\label{appendix:baselines}

As stated in Section \ref{section:baselines}, we compare RM-Distiller against several reward model (RM) distillation baselines:

\begin{itemize}
\item \textbf{Margin BT} \cite{touvron2023llama}: This approach extends the standard BT model by introducing a margin $\gamma$ to the loss function, aiming to improve the separability of chosen and rejected responses:

\begin{equation}
    \mathcal{L}_{\text{MBT}} = -\log \sigma (r_\phi(x, y_w) - r_\phi(x, y_l) - \gamma(x,y))
\end{equation}
where $\gamma(x,y) = s_T(x, y_w) - s_T(x, y_l)$ is derived from the teacher-assigned scores in our experiments.

\item \textbf{SteerLM} \cite{dong-etal-2023-steerlm}: Instead of pairwise ranking, SteerLM treats reward modeling as a regression task. The model is trained to predict specific quality dimensions based on teacher-assigned scalar values $s_i$:

\begin{equation}
\mathcal{L}_{\text{SteerLM}}
= \sum_i \left( r_{\phi,i}(x,y) - s_{T,i} \right)^2
\end{equation}

In our experiments, we simplify this into a single-attribute regression task by utilizing a global quality score provided by the teacher model, which can actually yield more superior performance \cite{ICLR2025_8e237ec6}.

\item \textbf{SynRM} \cite{ye-etal-2025-improving}: This method utilizes critiques generated by the teacher LLM. During training and evaluation, the critique $c$ is concatenated with the prompt and response, and the model is trained to align its scalar output with the teacher's preference while conditioned on the reasoning trace.

\begin{table}[!t]
\centering
\resizebox{0.7\linewidth}{!}{
\begin{tabular}{lc}
\toprule
Hyperparameter & RM \\
\midrule
Epochs              & 1 \\
Batch size                   & 16 \\
Optimizer                    & AdamW \\
Learning rate                & 1.00E--05 \\
Learning rate scheduler      & Cosine \\
Maximum sequence length      & 1024 \\
\bottomrule
\end{tabular}
}
\caption{Hyperparameter settings for student RM training.}
\label{tab:hyperparams_rm}
\end{table}

\item \textbf{RMBoost} \cite{shen2025rmboost}: This method focuses on data augmentation where the teacher model is used to improve or degrade its own responses by incorporating various quality factors into the prompts.

\item \textbf{CLoud} \cite{ankner2024critiqueoutloud}:  This approach explicitly distills reasoning capabilities into the RM through a two-stage process. First, the student model is finetuned using critiques generated by the teacher model. Second, a reward head is integrated with the language model backbone, enabling the reward model generating a critique via the language head before producing a final preference score via the reward head.

\item \textbf{LSAM} \cite{wang-etal-2024-reward-modeling}: This method applies adaptive margin-based label smoothing for RM training. Instead of assigning hard binary preference labels, LSAM introduces a soft target $\alpha(x, y)$ that reflects the confidence of each preference pair. The reward model is trained using the following objective:

\noindent
\begin{equation}
\resizebox{0.91\linewidth}{!}{%
$\displaystyle
\begin{aligned}
\mathcal{L}_{\text{LSAM}}(r_\psi) = & - \epsilon(x,y)\log p_\psi(y_w \succ y_l \mid x) \\
& - \big(1-\epsilon(x,y)\big) \log\!\big(1 - p_\psi(y_w \succ y_l \mid x)\big)
\end{aligned}$%
}
\end{equation}

\noindent

where $p_{\psi}(y_w \succ y_l \mid x) = \sigma(r_{\psi}(x, y_w) - r_{\psi}(x, y_l))$ denotes the model's predicted preference probability, and $\epsilon(x,y) = \sigma(s_T(x, y_w) - s_T(x, y_l))$ is the target soft label derived from the teacher's scores. This adaptive margin softens supervision for ambiguous preferences while preserving strong signals for confident comparisons.
\end{itemize}

\section{Additional Experiments}
\label{appendix:experiment}

\begin{table*}[!t]
\centering
\resizebox{0.75\linewidth}{!}{
\begin{tabular}{lccc}
\toprule
Hyperparameter & PPO & GRPO & DAPO \\
\midrule
data.gen\_batch\_size                                         & 256      & 256      & 512 \\
data.train\_batch\_size                                       & 256      & 256      & 256 \\
data.max\_prompt\_length                                      & 1024     & 1024     & 1024 \\
data.max\_response\_length                                    & 4096     & 4096     & 4096 \\
actor\_rollout\_ref.actor.optim.lr                            & 3.00E--06 & 3.00E--06 & 3.00E--06 \\
actor\_rollout\_ref.actor.ppo\_mini\_batch\_size              & 64       & 64       & 64 \\
actor\_rollout\_ref.actor.ppo\_micro\_batch\_size\_per\_gpu   & 8        & 8        & 8 \\
actor\_rollout\_ref.actor.use\_kl\_loss                       & false    & false    & false \\
actor\_rollout\_ref.rollout.log\_prob\_micro\_batch\_size\_per\_gpu & 16 & 16 & 16 \\
actor\_rollout\_ref.rollout.n                                 & —        & 8        & 8 \\
actor\_rollout\_ref.ref.log\_prob\_micro\_batch\_size\_per\_gpu & 16      & 16       & 16 \\
critic.optim.lr                                               & 1.00E--05 & —        & — \\
critic.ppo\_micro\_batch\_size\_per\_gpu                      & 8        & —        & — \\
trainer.n\_gpus\_per\_node                                    & 8        & 8        & 8 \\
trainer.nnodes                                                & 2        & 2        & 2 \\
trainer.total\_epochs                                         & 10       & 10       & 10 \\
\bottomrule
\end{tabular}
}
\caption{Hyperparameter settings for RL experiments.}
\label{tab:hyperparams_rl}
\end{table*}
\begin{table*}[!t]
\resizebox{\textwidth}{!}{
\begin{tabular}{lllrrrrrrrrr}
\toprule
\multirow{2}{*}{Student} & \multirow{2}{*}{Teacher} & \multirow{2}{*}{Method}
& \multicolumn{5}{c}{RewardBench} & \multicolumn{4}{c}{RM-Bench} \\
\cmidrule(lr){4-8} \cmidrule(lr){9-12}
& & & Chat & Hard & Safety & Reason & Avg.
& Easy & Normal & Hard & Avg. \\
\midrule

\multirow{2}{*}{Qwen2.5-3B-Instruct} 
& \multirow{2}{*}{Qwen3-14B} 
& Pairwise Scoring                     & \textbf{94.1} & 43.6 & 76.4 & \textbf{88.9} & 75.8 
& \textbf{90.2} & 64.8 & 27.1 & 60.7 \\
& & Self-Calibrated Scoring           & 93.3 & \textbf{54.8} & \textbf{81.2} & 85.4 & \textbf{78.7}
& 88.8 &\textbf{65.4} & \textbf{30.3} & \textbf{61.5} \\
\bottomrule
\end{tabular}
}
\centering
\caption{Results of ablation study on Self-Calibrated Scoring.}
\label{tab:ablation_cali}
\end{table*}

\begin{table*}[!t]
\resizebox{0.9\textwidth}{!}{
\begin{tabular}{lllrrrrrrrrr}
\toprule
\multirow{2}{*}{Student} & \multirow{2}{*}{Teacher} & \multirow{2}{*}{Weight}
& \multicolumn{5}{c}{RewardBench} & \multicolumn{4}{c}{RM-Bench} \\
\cmidrule(lr){4-8} \cmidrule(lr){9-12}
& & & Chat & Hard & Safety & Reason & Avg.
& Easy & Normal & Hard & Avg. \\
\midrule
\multirow{8}{*}{Qwen2.5-3B-Instruct} 
& \multirow{4}{*}{GPT-4o} 
& $\alpha{=}0.0$ & 94.1 & 43.6 & 80.1 & 89.1 & 76.8
& 79.8 & 61.8 & 34.6 & 58.8 \\
& & $\alpha{=}0.1$  & 94.1 & 48.7 & 82.7 & 87.3 & 78.2 
& 80.6 & 63.1 & \textbf{37.7} & 60.5 \\
& & $\alpha{=}0.2$  & \textbf{95.5} & \textbf{51.1} & \textbf{84.1} & \textbf{87.4} & \textbf{79.5} 
& 86.4 & \textbf{64.8} & 32.2 & \textbf{61.1} \\
& & $\alpha{=}0.3$  & 93.9 & 46.5 & 79.1 & 84.6 & 76.0 
& \textbf{87.1} & 61.5 & 27.7 & 58.8 \\
\cmidrule{2-12}
& \multirow{4}{*}{Qwen3-14B} 
& $\beta{=}0.0$   & 93.6 & 53.1 & 80.3 & 88.8 & 78.9 
& 89.1 & 64.5 & 27.5 & 60.4 \\
& & $\beta{=}0.1$   & 93.6 & 53.3 & 81.6 & 84.4 & 78.2 
& \textbf{89.5} & \textbf{66.7} & 30.4 & 62.2 \\
& & $\beta{=}0.2$   & \textbf{95.0} & \textbf{55.0} & \textbf{81.8} & \textbf{89.0} & \textbf{80.2}
& 87.2 & 66.3 & \textbf{34.6} & \textbf{62.7} \\
& & $\beta{=}0.3$   & 94.7 & 54.6 & 81.5 & 85.0 & 79.0 
& 88.0 & 65.3 & 31.5 & 61.6 \\
\bottomrule
\end{tabular}
}
\centering
\caption{Results of different regularization weights.}
\label{tab:ablation_weight}
\end{table*}

\begin{table}[!t]
\centering
\begin{tabular}{crrrrr}
\toprule
\multirow{2}{*}{\makecell{Margin\\Threshold}} & \multicolumn{5}{c}{RewardBench} \\
\cmidrule(lr){2-6}
 & Chat & Hard & Safety & Reason & Avg. \\
\midrule
0 & 95.0 & 52.4 & 79.1 & 81.2 & 79.2 \\
1 & 93.9 & 52.9 & 82.4 & 86.9 & 79.0 \\
2 & 95.0 & 54.2 & 81.2 & 89.1 & 79.9 \\
3 & \textbf{95.0} & \textbf{55.0} & 81.8 & 89.0 & \textbf{80.2} \\
4 & 94.1 & 52.6 & \textbf{83.7} & \textbf{89.2} & 79.9 \\
\bottomrule
\end{tabular}
\caption{Results of different score margin thresholds.}
\label{tab:ablation_margin}
\end{table}

\begin{table}[!t]
\centering
\begin{tabular}{llr}
\toprule
\textbf{Policy} & \textbf{Method} & \textbf{Score} \\
\midrule
\multirow{4}{*}{Qwen-3-8B} & Baseline & 3.1 \\
                           & BT Classifier & 3.4 \\
                           & RM-Distiller & 3.5 \\
                           & RM-Distiller + Rule & \textbf{3.6} \\
\bottomrule
\end{tabular}
\caption{Results in the E-commerce customer service scenario.}
\label{tab:ecommerce_results}
\end{table}

\begin{table*}[!t]
\resizebox{\textwidth}{!}{
\begin{tabular}{lllrrrrrrrrr}
\toprule
\multirow{2}{*}{Student} & \multirow{2}{*}{Teacher} & \multirow{2}{*}{Method}
& \multicolumn{5}{c}{RewardBench} & \multicolumn{4}{c}{RM-Bench} \\
\cmidrule(lr){4-8} \cmidrule(lr){9-12}
& & & Chat & Hard & Safety & Reason & Avg.
& Easy & Normal & Hard & Avg. \\
\midrule
\multirow{4}{*}{Qwen2.5-3B-Instruct} 
& \multirow{4}{*}{GPT-4o} 
& BT Classifier                      & 93.3 & 37.7 & 69.3 & 81.0 & 70.3 
& \textbf{91.9} & 62.3 & 13.6 & 55.9 \\
& & + Margin-Aware Regression          & 94.7 & 41.2 & 75.7 & 79.2 & 72.7 
& 90.9 & 61.9 & 19.0 & 57.3 \\
& & + Contrastive Refinement           & 94.1 & 43.6 & 80.1 & \textbf{89.1} & 76.8
& 79.8 & 61.8 & \textbf{34.6} & 58.8 \\
& & + Generative Regularization         & \textbf{95.5} & \textbf{51.1} &\textbf{84.1}  & 87.4 & \textbf{79.5}
& 86.4 & \textbf{64.8} & 32.2 & \textbf{61.1} \\
\midrule
\multirow{4}{*}{Qwen2.5-3B-Instruct} 
& \multirow{4}{*}{Qwen3-14B} 
& BT Classifier                      & 88.0 & 43.4 & 72.4 & 83.5 & 71.8 
& \textbf{89.8} & 60.7 & 19.8 & 56.7 \\
& & + Margin-Aware Regression          & 95.0 & 45.0 & 76.9 & 86.0 & 75.7 
& 89.6 & 63.4 & 25.3 & 59.5 \\
& & + Contrastive Refinement           & 93.3 & 54.8 & 81.2 & 85.4 & 78.7
& 88.8 & 65.4 & 30.3 & 61.5 \\
& & + Generative Regularization          & \textbf{95.0} & \textbf{55.0} & \textbf{81.8} & \textbf{89.0} & \textbf{80.2}
& 87.2 & \textbf{66.3} & \textbf{34.6} & \textbf{62.7} \\
\bottomrule
\end{tabular}
}
\centering
\caption{Detailed results of ablation study on RM-Distiller components.}
\label{tab:ablation_full}
\end{table*}

\begin{table*}[!ht]
\resizebox{\linewidth}{!}{
\begin{tabular}{lllrrrrrrrrr}
\toprule
\multirow{2}{*}{Student} & \multirow{2}{*}{Teacher} & \multirow{2}{*}{Method}
& \multicolumn{5}{c}{RewardBench} & \multicolumn{4}{c}{RM-Bench} \\
\cmidrule(lr){4-8} \cmidrule(lr){9-12}
& & & Chat & Hard & Safety & Reason & Avg.
& Easy & Normal & Hard & Avg. \\
\midrule
\multirow{4}{*}{Qwen2.5-3B-Instruct} 
& \multirow{2}{*}{GPT-4o} 
& BT Classifier & 77.7 & 33.1 & 41.5 & 48.7 & 50.3 
& \textbf{86.2} & 47.8 & 10.3 & 48.1 \\
& & \textbf{RM-Distiller}      & \textbf{94.1} & \textbf{57.7} & \textbf{83.8} & \textbf{68.6} & \textbf{76.1} 
& 85.7 & \textbf{62.6} & \textbf{31.4} & \textbf{59.9} \\
\cmidrule{2-12}

& \multirow{2}{*}{Qwen3-14B} 
& BT Classifier & 82.4 & 52.0 & 56.0 & 67.9 & 64.6 
& 86.5 & 54.2 & 15.2 & 51.9 \\
& & \textbf{RM-Distiller}      & \textbf{89.9} & \textbf{57.2} & \textbf{77.8} & \textbf{80.7} & \textbf{76.4} 
& \textbf{87.6} & \textbf{61.2} & \textbf{31.9} & \textbf{60.2} \\
\bottomrule
\end{tabular}}
\caption{Detailed results of different methods under 1K instructions.}

\label{tab:data_efficiency_full}
\end{table*}

\begin{table*}[!ht]
\centering
\resizebox{\textwidth}{!}{
\begin{tabular}{lllrrrrrrr}
\toprule
\multirow{2}{*}{Student} & \multirow{2}{*}{Teacher} & \multirow{2}{*}{Method} & \multicolumn{7}{c}{EvalBiasBench} \\
\cmidrule(lr){4-10}
& & & Length & Concreteness & \makecell{Empty \\ Reference} & \makecell{Content \\ Continuation} & \makecell{Nested \\ Instruction} & \makecell{Familiar \\ Knowledge} & Overall \\
\midrule
\multicolumn{3}{l}{GPT-4o \cite{hurst2024gpt}} & 44.1 & 100.0 & 65.4 & 95.8 & 41.7 & 79.2 & 70.0 \\
\multicolumn{3}{l}{Qwen3-14B \cite{yang2025qwen3}} & 47.1 & 96.4 & 92.3 & 95.8 & 58.3 & 66.7 & 75.0 \\
\multicolumn{3}{l}{Qwen2.5-3B-Instruct \cite{qwen2024qwen25}} & 20.6 & 67.9 & 42.3 & 54.2 & 33.3 & 54.2 & 44.4 \\
\midrule
\multirow{4}{*}{Qwen2.5-3B-Instruct} & \multirow{2}{*}{GPT-4o} & BT Classifier & 5.9 & \textbf{50.0} & 7.7 & 66.7 & 33.3 & 25.0 & 30.0 \\
& & \textbf{RM-Distiller} & \textbf{17.7} & 42.9 & \textbf{69.2} & \textbf{100.0} & \textbf{58.3} & \textbf{75.0} & \textbf{57.5} \\
\cmidrule{2-10}
& \multirow{2}{*}{Qwen3-14B} & BT Classifier & 11.8 & 21.4 & \textbf{100.0} & 50.0 & 41.7 & \textbf{75.0} & 47.5 \\
& & \textbf{RM-Distiller} & \textbf{29.4} & \textbf{71.4} & 84.6 & \textbf{100.0} & \textbf{58.3} & 50.0 & \textbf{63.8} \\
\bottomrule
\end{tabular}
}
\caption{Detailed results of different methods on EvalBiasBench.}
\label{tab:evalbiasbench_full}
\end{table*}

\subsection{Ablation of Self-Calibrated Scoring}
As described in Section \ref{sec:cali}, we introduce Self-Calibrated Scoring to assign more precise scores to the refined preference pairs in Contrastive Refinement. To evaluate its effectiveness, we conduct a comparative analysis between our self-calibrated approach and pairwise scoring baseline, where the teacher LLM provides scores for the $(y_w^*, y_l)$ pairs without the reference score and refinement history. In both settings, the resulting scores are utilized to train the student RM via Margin-Aware Regression.

As shown in Table \ref{tab:ablation_cali}, Self-Calibrated Scoring significantly outperforms the pairwise scoring baseline. We observe that directly scoring the pair $(y_w^*, y_l)$ often yields negligible margins or ties, as teacher models struggle to distinguish subtle improvements between highly correlated responses in a zero-shot setting. In contrast, by incorporating the score of rejected response as an anchor and leveraging the refinement history, Self-Calibrated Scoring produces more discriminative score differences. These precise signals provide the student with clearer optimization targets, which are essential for capturing fine-grained quality boundaries.

\subsection{Ablation of the Regularization Weights}
\label{appendix:weight}
As stated in Section \ref{section:generation}, during Generative Regularization, we apply two regularization weights $\alpha$ and $\beta$ for negative log-likelihood (NLL) loss and Kullback–Leibler (KL) loss, respectively. In this section, we analyze the impact of the two weights on the distillation performance. Given that GPT-4o is a closed-source model that does not provide access to token-level log-probabilities, direct computation of KL divergence is unfeasible. Therefore, we employ a two-stage ablation strategy:
\begin{enumerate}[itemsep=1mm, parsep=0pt]
    \item NLL weight $\alpha$: With GPT-4o as the teacher model, we set $\beta = 0$ and evaluate the student's performance across different values of $\alpha$.

    \item KL weight $\beta$: With Qwen3-14B as the teacher model, we set the optimal $\alpha$ from the last stage and evaluate the student's performance across different values of $\beta$.
\end{enumerate}

As shown in Table \ref{tab:ablation_weight}, incorporating Generative Regularization helps prevent the model from overfitting to superficial preference patterns and maintains its generalization across diverse tasks. Moreover, our empirical observations indicate that optimal performance is achieved when both $\alpha$ and $\beta$ are set to 0.2, providing the best balance between discriminative accuracy and linguistic consistency.

\subsection{Ablation of the Filtering Threshold}
\label{appendix:threshold}
As stated in Section \ref{sec:filtering}, we leverage two criteria for filtering. For the edit distance threshold $\tau_e$, we set it to 0, which means we only filter out samples where the teacher model makes no modifications to the rejected response. For the score margin threshold $\tau_s$, which determines the minimal score difference between preference pairs, in this section, we investigate its impact during distillation from Qwen3-14B.

As shown in Table \ref{tab:ablation_margin}, the choice of $\tau_s$ significantly affects the quality of the distilled reward model. When the threshold is too low, the model's performance drops, likely due to the inclusion of low-quality refinements where the teacher's modifications do not lead to a sufficient improvement in response quality. Conversely, an excessively high threshold also leads to a performance decline. This may be because a large threshold filters out subtler preference pairs, preventing the student from learning to distinguish fine-grained quality differences that are essential for reliable alignment. Empirically, we find that a balanced threshold of $\tau_s = 3$ yields the best results, as it effectively filters out noise while preserving the informative preference boundaries.

\subsection{Statistical Indicators of RL}
\begin{figure*}[!t]
    \centering
    \includegraphics[width=0.98\linewidth]{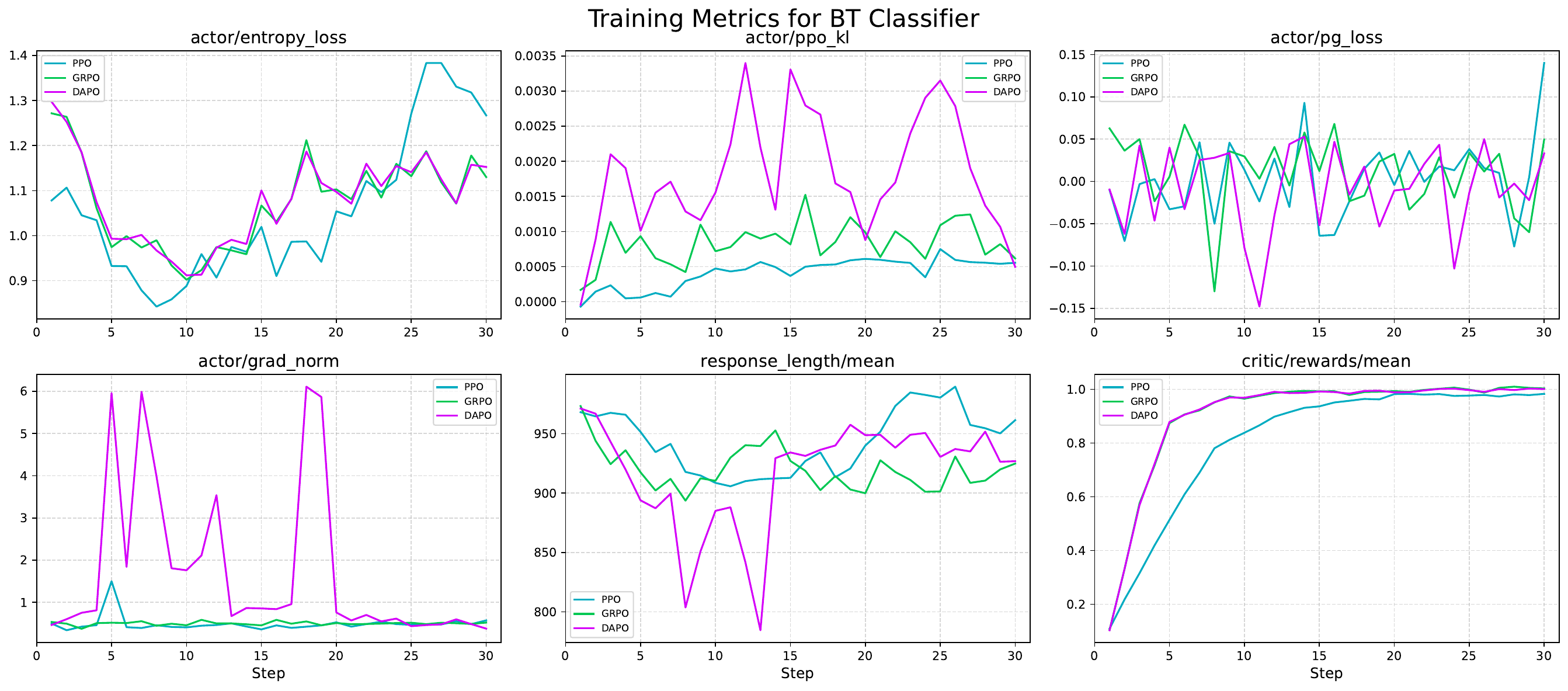}
    \caption{The variation of statistical metrics on ShareGPT using the BT Classifier across different RL algorithms.}
    \label{figure:rlhf-bt}
\end{figure*}

\begin{figure*}[!t]
    \centering
    \includegraphics[width=0.98\linewidth]{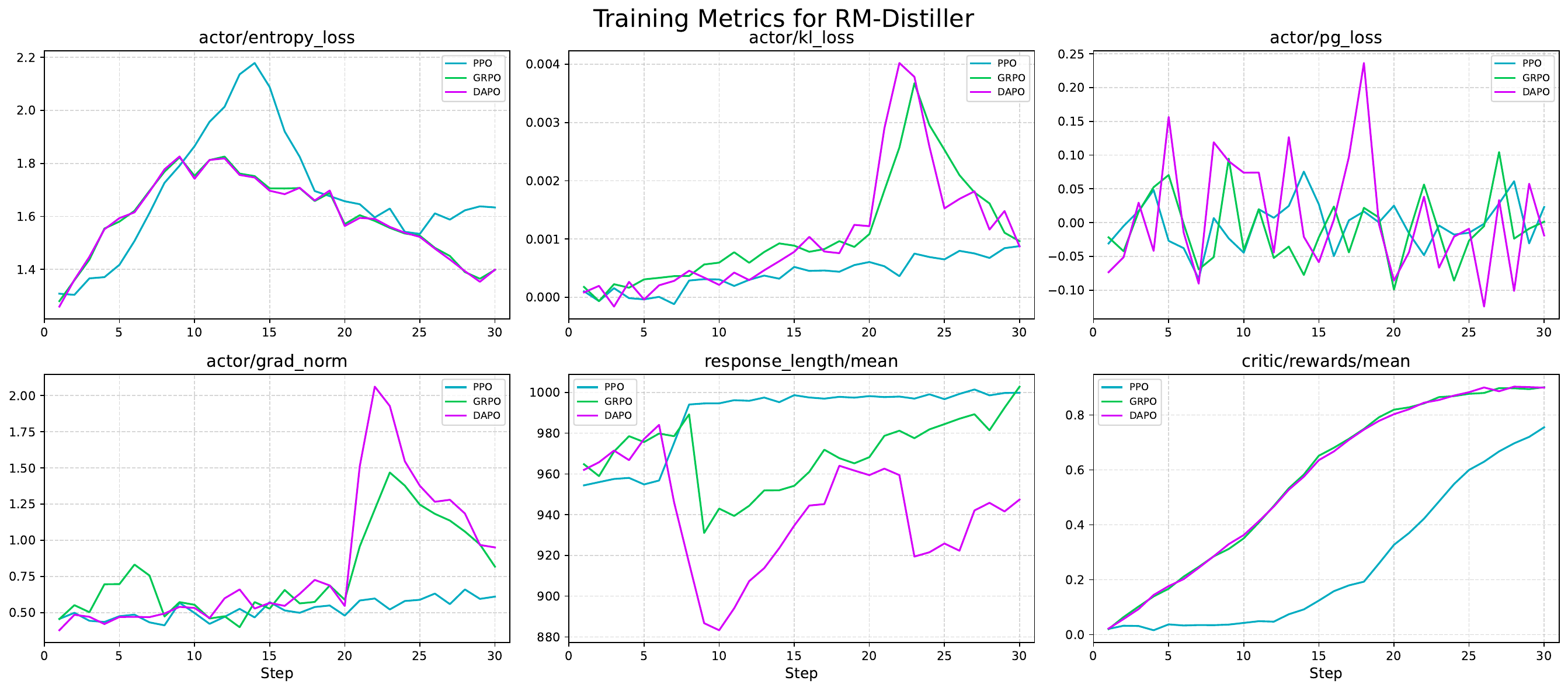}
    \caption{The variation of statistical metrics on ShareGPT using the RM-Distiller across different RL algorithms.}
    \label{figure:rlhf-distil}
\end{figure*}

As stated in Section \ref{sec:rlhf}, to verify the effectiveness of our RM distillation method in real application, we compare different RMs by applying them to RL training. Figures \ref{figure:rlhf-bt} and \ref{figure:rlhf-distil} display the training dynamics of PPO, GRPO, and DAPO on ShareGPT based on the BT Classifier and RM-Distiller, respectively. As can be seen, despite the variation of RL algorithms we have used, RM-Distiller demonstrates more stable training dynamics compared with the BT Classifier, with smoother gradient norms and reduced fluctuations in policy-related metrics. These trends indicate that distilled preference signals provide more reliable supervision during RL.

\subsection{RL Performance in Specialized Application}
Building on the RLHF experiments in Section \ref{sec:rlhf}, to further validate the effectiveness in real-world applications, we evaluate RM-Distiller in specialized business context, specifically in an E-commerce customer service scenario. In this domain, LLMs act as customer agents, necessitating strict adherence to both communicative quality and rule-based constraints. Specifically, we reuse the RMs trained in Section \ref{sec:rlhf}, and utilize Qwen-3-8B \cite{yang2025qwen3} as the base policy and conduct reinforcement learning using the DAPO algorithm. For evaluation, five expert annotators independently scored model responses on a scale of 1 to 5 across 1,000 internal test samples. We compare the average scores of policies aligned with different reward signals in Table \ref{tab:ecommerce_results}.

As can be seen, policy model supervised by RMs distilled from RM-Distiller outperforms model supervised BT Classifier by a large margin in this specific vertical application, indicating its superior capability to capture preference signals. Furthermore, integrating rule-based rewards with RM-Distiller yields the highest overall satisfaction score. Overall, RM-Distiller improves the model's performance in specific vertical applications, which further validates the effectiveness and generalizability of our RM distillation method.

\section{Detailed Experiment Results}
\label{appendix:detail_results}

In this section, we present detailed experiment results which are omitted in the main body of this paper due to space limitation. Table \ref{tab:ablation_full} presents the full per-category results of the ablation study on RM-Distiller components across RewardBench and RM-Bench. Table \ref{tab:data_efficiency_full} reports the full per-category results of the BT Classifier and RM-Distiller on RewardBench and RM-Bench under the 1K-instruction setting. Table \ref{tab:evalbiasbench_full} provides the detailed results of performance on EvalBiasBench across various bias categories.

\section{Prompt Templates}
\label{appendix:prompt}
In this section, we provide the prompt templates used for both the reproduced baselines and our proposed RM-Distiller. The prompts for Initial Preference Annotation, Contrastive Refinement, and Self-Calibrated Scoring are presented in Prompt \ref{prompt:prompt_score}, \ref{prompt:prompt_refine}, and \ref{prompt:prompt_cali_score}. Additionally, the prompt templates for critique generation (used in SynRM and CLoud) and response rewriting (used in RMBoost) are presented in Prompt \ref{prompt:prompt_synrm} and \ref{prompt:prompt_rmboost}.
\begin{figure*}[!t]
\begin{promptbox}[title={Prompt~\thetcbcounter: Prompt Template for Initial Preference Annotation}, label={prompt:prompt_score}]

You are a helpful and precise assistant for checking the quality of the answer. \\

[Question] \\
\textbf{\{question\_body\}} \\

[The Start of Assistant 1's Answer] \\
\textbf{\{answer1\_body\}} \\
{} [The End of Assistant 1's Answer] \\

[The Start of Assistant 2's Answer] \\
\textbf{\{answer2\_body\}} \\
{} [The End of Assistant 2's Answer] \\

[System] \\
We would like to request your feedback on the performance of two AI assistants in response to the user question displayed above. Please rate the helpfulness, relevance, accuracy, level of details of their responses.
Each assistant receives an overall score on a scale of 1 to 10, where a higher score indicates better overall performance.
Please output a single line containing only two values indicating the scores for Assistant 1 and 2, respectively. The two scores are separated by a space. Avoid any potential bias and ensure that the order in which the responses were presented does not affect your judgment. Do NOT provide any explanation. \\

\#\#\# Response:
\end{promptbox}

\end{figure*}

\begin{figure*}[!t]
\begin{promptbox}[title={Prompt~\thetcbcounter: Prompt Template for Contrastive Refinement}, label={prompt:prompt_refine}]

<Start of Instruction> \\
Given the question from the user, you need to modify the given non-preferred to get a preferred reply from the assistant. Begin your answer with why the given non-preferred response is not preferred compared to the preferred response and how this can be improved. Output your editing by strictly following this format: \\

<Start of Reasoning> \\
<End of Reasoning> \\

<Start of Modified Response> \\
<End of Modified Response> \\

You must maintain the original structure of the non-preferred response, ensure the edit distance between the modified response and the non-preferred response is as low as possible. \\
<End of Instruction> \\

<Start of Question> \\
\textbf{\{question\_body\}} \\
<End of Question> \\

<Start of Preferred Response> \\
\textbf{\{teacher\_answer\_body\}} \\
<End of Preferred Response> \\

<Start of Non-Preferred Response> \\
\textbf{\{rejected\_answer\_body\}} \\
<End of Non-Preferred Response>

\end{promptbox}
\end{figure*}

\begin{figure*}[!t]
\begin{promptbox}[title={Prompt~\thetcbcounter: Prompt Template for Self-Calibrated Scoring}, label={prompt:prompt_cali_score}]

\textbf{\{refinement\_history\}} \\ \\

Now, considering the rejected response was scored \textbf{\{rejected\_score\}} out of 10 for quality, helpfulness, and alignment, score the Modified Response in the assistant's reply above on the same 1-10 scale. Output only the numerical score. Do NOT provide any explanation.

\end{promptbox}
\end{figure*}

\begin{figure*}[!t]
\begin{promptbox}[title={Prompt~\thetcbcounter: Prompt Template for Critique Generation of SynRM and CLoud}, label={prompt:prompt_synrm}]

User: \textbf{\{question\_body\}} \\
Chatbot: \textbf{\{answer\_body\}} \\
Please provide a critique of the last response in one short paragraph. Your critique should be concise, specific, insightful and to the point. Aspects you should consider are: (1) Helpfulness. A good response should precisely/closely answer the user's request. (2) Correctness. A good response should be honest and factually correct.

\end{promptbox}

\end{figure*}
\begin{figure*}[!t]
\begin{promptbox}[title={Prompt~\thetcbcounter: Prompt Template for Response Rewriting of RMBoost}, label={prompt:prompt_rmboost}]

Your task is to generate another response by EDITING the given response, so that the new response is \textbf{\{preference\}} than the given response with respect to some evaluation aspects. \\

<task\_description> \\
Below you will first see a guideline with detailed evaluation aspects of responses. \\
Then, you will be presented with the question and the given response. \\

You should complete the following steps internally: \\
Step 1: Select a few aspects from the guideline. \\
Step 2: Generate another response that is \textbf{\{preference\}} than the given response in terms of above selected aspects. \\
</task\_description> \\

<guideline> \\
A high-quality response should: \\
- Directly and correctly answer the question. \\
- Be clear, coherent, and logically well-structured. \\
- Be factually and logically consistent. \\
- Include all necessary reasoning steps or explanations. \\

We evaluate responses from the following aspects: \\
- (Honesty): The assistant should be honest about whether it knows the answer and express its uncertainty explicitly. Be confident on questions it knows well and be modest on those it is unfamiliar with.  \\
- (Truthfulness): The assistant should answer truthfully and be faithful to factual knowledge as well as given contexts, never making up any new facts that are not true or cannot be grounded in the instruction. \\
- (Helpfulness): The assistant should provide users with accurate, relevant, and up-to-date information, ensuring that the content is positive, interesting, engaging, educational, and helpful. \\
- (Relevance and Coherence): Whether the response is focused on the question, logically organized, and free of irrelevant or confusing content. \\
- (Correctness and Faithfulness): Whether the response contains correct reasoning and does not include factual or logical errors. \\
- (Completeness): Whether the response includes all necessary reasoning steps needed to fully answer the question. \\
- (Safety and Ethics): Whether the response adheres to ethical guidelines, avoids harmful or biased content, and respects user privacy and dignity. \\
</guideline> \\

Below is the question. \\
<question> \\
\textbf{\{question\_body\}} \\
</question> \\

Below is the given response. \\
<given\_response> \\
\textbf{\{answer\_body\}} \\
</given\_response> \\

Read the question and given response carefully. Review the above task description and guideline. Think about how to accomplish the task step by step before you reply. Put your generated response in <response></response> tags.

\end{promptbox}
\end{figure*}

\end{CJK*}
\end{document}